# Autonomous Exploration-Based Precise Mapping for Mobile Robots through Stepwise and Consistent Motions

Muhua Zhang[1], Lei Ma[1], Ying Wu[1], Kai Shen[1], Yongkui Sun[1] and Henry Leung[2], *Fellow, IEEE*

***Abstract*—This paper presents an autonomous exploration framework. It is designed for indoor ground mobile robots that utilize laser Simultaneous Localization and Mapping (SLAM), ensuring process completeness and precise mapping results. For frontier search, the local-global sampling architecture based on multiple Rapidly Exploring Random Trees (RRTs) is employed. Traversability checks during RRT expansion and global RRT pruning upon map updates eliminate unreachable frontiers, reducing potential collisions and deadlocks. Adaptive sampling density adjustments, informed by obstacle distribution, enhance exploration coverage potential. For frontier point navigation, a stepwise consistent motion strategy is adopted, wherein the robot strictly drives straight on approximately equidistant line segments in the polyline path and rotates in place at segment junctions. This simplified, decoupled motion pattern improves scan-matching stability and mitigates map drift. For process control, the framework serializes frontier point selection and navigation, avoiding oscillation caused by frequent goal changes in conventional parallelized processes. The waypoint retracing mechanism is introduced to generate repeated observations, triggering loop closure detection and backend optimization in graph-based SLAM, thereby improving map consistency and precision. Experiments in both simulation and real-world scenarios validate the effectiveness of the framework. It achieves improved mapping coverage and precision in more challenging environments compared to baseline 2D exploration algorithms. It also shows robustness in supporting resource-constrained robot platforms and maintaining mapping consistency across various LiDAR field-of-view (FoV) configurations.**

## I. Introduction

Autonomous exploration-based mapping enables mobile robots to navigate unknown environments and construct maps without human intervention. **Figure 1** highlights its value for traditional laser Simultaneous Localization and Mapping (SLAM)-based inspection and service mobile robots [1]-[4] as well as embodied intelligent mobile robots [5]-[7] which still depend on localization and path planning using grid maps. For the former, conventional deployments often involve manual mapping processes performed by experienced robot operators, which are intensive and labor-demanding. The introduction of autonomous exploration-based mapping reduces deployment costs for these robots. For the latter, advancements in autonomous exploration-based mapping can enhance the robot's foundational environmental perception capabilities,

This work was supported by the National Natural Science Foundation of China (Grants 62203371, 61733015, U1934204). (Corresponding author: Kai Shen). Full HD video at `https://youtu.be/Lw-3I2e544I`.

[1] School of Electrical Engineering, Southwest Jiaotong University, Chengdu 611756, China (`{muhua.zhang@my., malei@, wy2022@my., shenkai@, yksun@}swjtu.edu.cn`).

[2] Department of Electrical and Computer Engineering, University of Calgary, Calgary AB T2N 1N4, Canada (`leungh@ucalgary.ca`).

Materials for effectiveness validations are available under request.

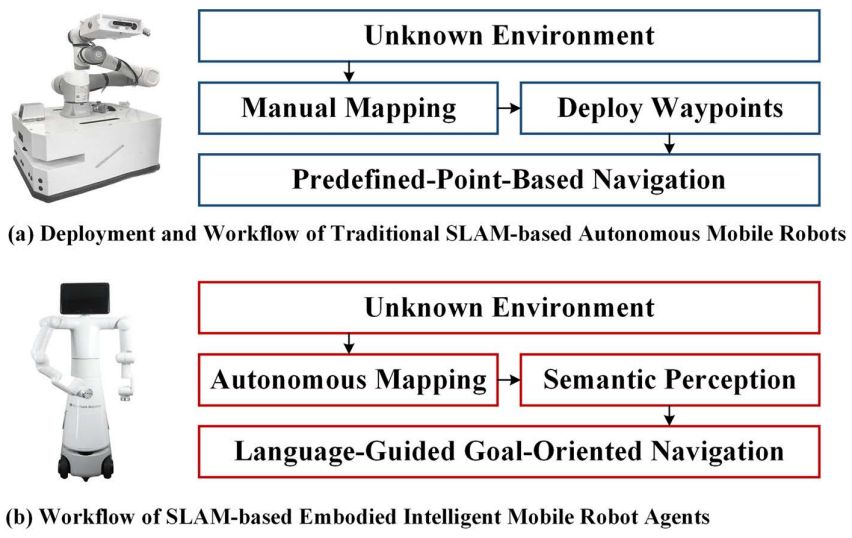

Fig. 1. Deployment and workflow of traditional and modern mobile robots.

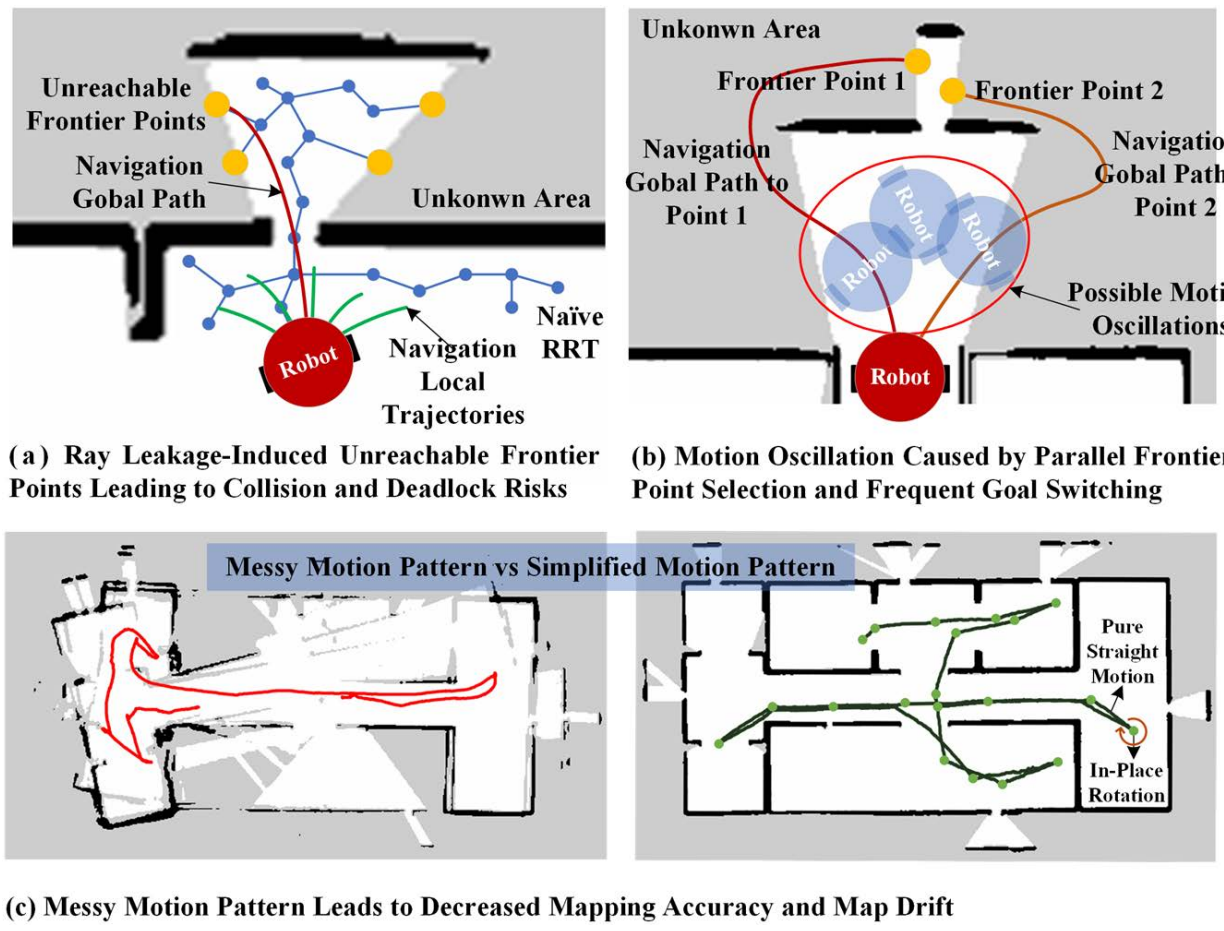

Fig. 2. Possible issues in autonomous exploration systems for mapping.

thereby better serving their upstream tasks. At the same time, high-quality maps are also crucial, especially for precise operations of robots. However, existing exploration systems still have challenges when operating in complex indoor environments and ensuring mapping precision, as illustrated in **Figure 2**. In **Figure 2(a)**, "ray leakage" refers to appearances of areas that can be scanned by the LiDAR but remain unreachable to the robot. In such cases, naive RRT with only edge-based collision check may expand frontier points into unreachable regions. Navigating these unreachable goals may lead to potential collisions or deadlocks, obstructing the exploration process. In **Figure 2(b)**, conventional parallel execution of frontier selection and navigation may cause frequent navigation goal switching, leading to oscillations and exploration stagnation. In **Figure 2(c)**, several exploration systems, for example [8]-[10], utilize general navigation frameworks like *move_base* [11] for moving, which generates messy motions. This results in SLAM scan matching having to estimate both rotation and translation terms simultaneously, which may introduce potential imprecisions of mapping [12], particularly on resource-constrained computing platforms or on field-of-view (FoV)-constrained LiDAR configurations.

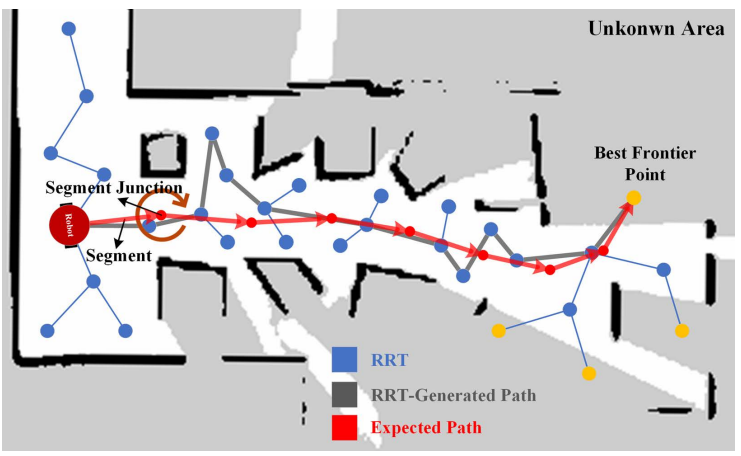

Fig. 3. Comparison of RRT-generated and expected polyline paths.

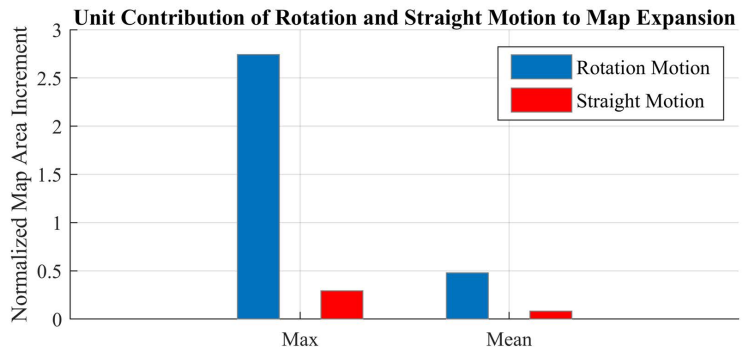

Fig. 4. Unit contribution of pure rotation and pure straight to map extension.

In contrast, simplified and decoupled motion patterns, such as pure linear motions and in-place rotations, enhance mapping stability and precision by minimizing trajectory-induced nonlinear errors and improving scan-matching consistency. The absence of a mechanism for retracing visited waypoints to generate repeated observations also limits the ability of existing exploration systems to fully leverage the potential of graph-based SLAM systems such as *Cartographer* [13], [14] and *SLAM Toolbox* [15]. Moreover, their lack of a guarantee of process completeness, such as determining the completion of exploration, further constrains their practical applicability.

To address the above challenges, this paper presents an autonomous exploration framework with guaranteed process completeness and precise mapping results for indoor SLAM mobile robots. It provides a complete closed-loop process, enabling the robot to start from a point in unknown, complex, and constrained environments, explore surroundings, perform precise mapping, determine completion, and autonomously return to the starting point. In this paper, precise mapping refers to constructing drift-free, mismatch-free maps that clearly represent the environment, without guaranteeing strict metric consistency with the physical world. In these maps, robots can reliably obtain highly repeatable pose estimations at the same point, meeting the requirements of indoor mobile robot precise operations. For frontier point search, it employs the local-global sampling architecture based on multiple RRT. Traversability checks during RRT expansion and global RRT pruning upon map updates eliminate unreachable frontiers, reducing risks of collisions and deadlocks. Additionally, adaptive sampling density adjustments based on obstacle distribution enhance exploration coverage potential. For frontier point navigation, as shown in **Figure 3**, it employs the stepwise and consistent motion pattern to improve mapping precision. The robot follows polyline paths composed of approximately equidistant line segments, moving along each segment and performing in-place rotations at junctions to align with the next segment. Maintaining uniform segment lengths ensures consistent environmental observation. **Figure 4,** derived from data analysis of a real mapping process using this simplified motion pattern, shows that in-place rotations contribute more to map expansion per unit time than linear motion. Therefore, while ensuring segment length uniformity, the cumulative in-place rotation angle distance should be maximized. **Figure 3** further illustrates that RRT-generated polyline paths suffer from uneven line segment lengths, penetration into narrow areas, and motion redundancy, making them unsuitable for frontier point navigations. To address these, a polyline path planner is proposed, integrating Dijkstra's algorithm as the frontend and incorporating both Bidirectional Interpolation (BI) [16], [17] and Dynamic Programming (DP) in the backend. This approach ensures segment uniformity while trying to increase the cumulative in-place rotation angle. For process control, the framework serializes frontier point selection and navigation, preventing oscillatory behaviors caused by frequent goal switching in conventional parallelized exploration and navigation. The visited waypoint retracing mechanism is also introduced, enabling more repeated observations to trigger loop closure detection and backend optimization in graph-based SLAM, thereby enhancing global mapping consistency and precision.

## II. Related Work

### A. Exploration in Unknown Environments

In recent years, autonomous exploration has been making continuous progress. Traditional exploration with grid map image-based frontier extraction [9], [18]-[20] uses computer vision algorithms to identify boundaries between known and unknown. However, these frontiers are derived purely from boundary information and do not ensure reachability. And as the map expands, computational costs increase. In contrast, RRT-based exploration [8], [10], [21]-[22] sample the space using randomly generated points, and random points in unknown areas are used as candidate frontier points, showing more advantages in efficiency, exploration coverage, and scalability. [8] and [23] propose accelerating exploration using multiple independently growing RRTs. [21] introduces a two-stage process that improves coverage through revisiting traversed areas. [10] and [22] refine RRT expansion strategies by introducing adaptive adjustment for sampling density or expansion boundary based on map data. [10] also examines the task allocation for multi-robot collaborative exploration. Additionally, various learning-based methods [24]-[27] have been applied to autonomous exploration. However, their generalization remains challenging. For example, the model proposed in [24] is trained in office environments and may struggle to adapt to other scenarios. Overall, existing exploration research primarily focuses on frontier search and multi-robot collaboration rather than precise mapping, often neglecting the impact of robot motion patterns. They frequently integrate traditional SLAM methods such as *gmapping* [28] but lack adaptation to modern graph-based SLAM. Moreover, they rarely evaluate consistency and robustness on resource-constrained computing platforms or across diverse LiDAR FoV configurations.

### B. Polyline Path Planning

Research on path polyline approximation, also known as path simplification, remains relatively limited. [29] proposes a key node extraction algorithm based on a divide-and-conquer approach, aiming to minimize deviations of simplified paths. *OMPL* [30] incorporates a path simplification algorithm that removes redundant points between two randomly selected path points within a predefined threshold window. [16] and [17] propose Bidirectional Interpolation (BI)-based path polyline approximation methods that focus on minimizing the number of segments in the path. Overall, these methods prioritize path simplification rather than guiding collision-free, stepwise, and consistent motions, making them unsuitable for direct integration into the proposed framework.

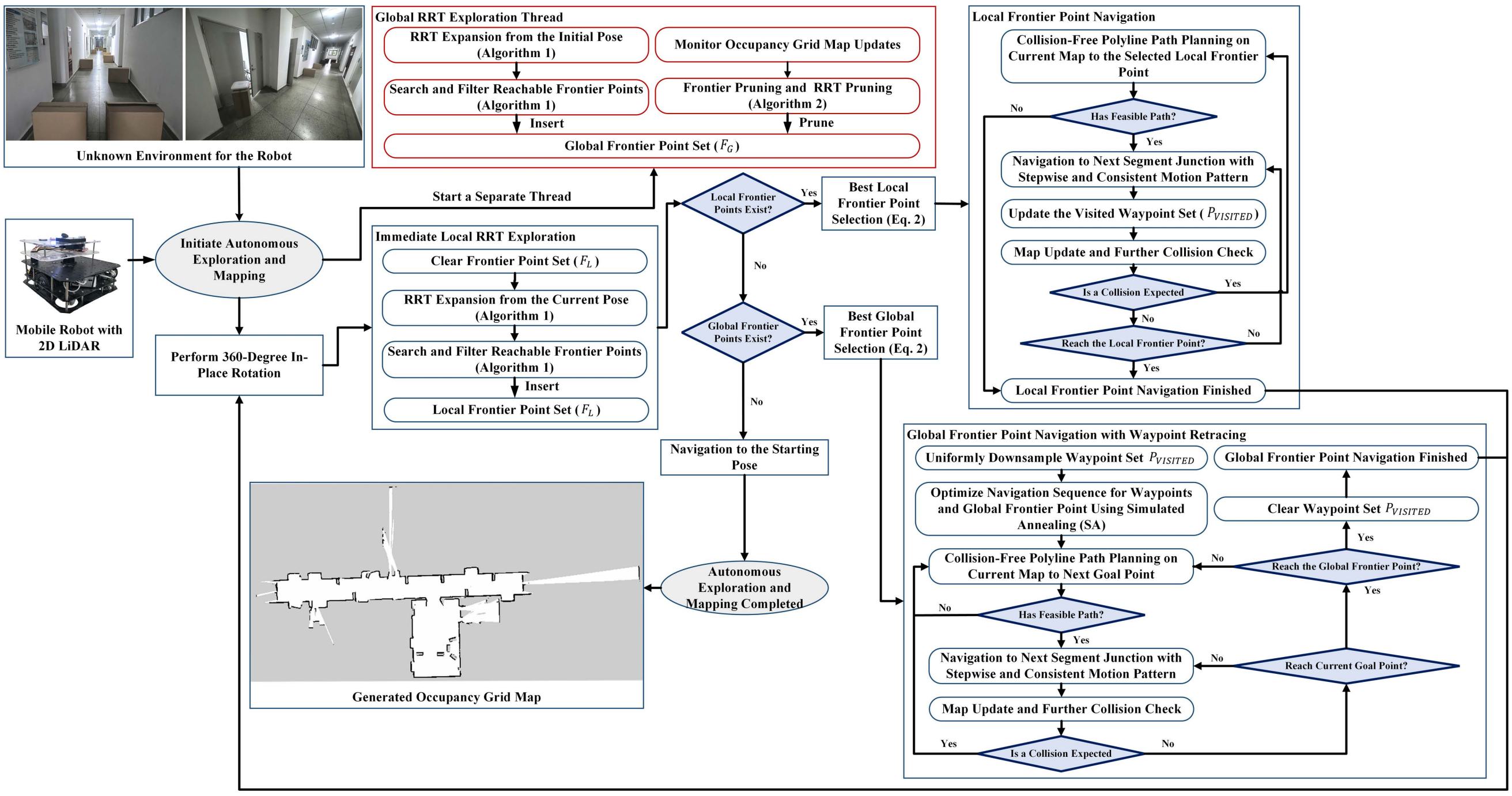

Fig. 5. The overall architecture and process design of the proposed framework.

## III. Contributions

This paper presents an autonomous exploration framework with guaranteed process completeness and precise mapping results. Experiments conducted in both simulation and real-world scenarios validate the framework's effectiveness in autonomous exploration-based mapping and its robustness on resource-constrained robot platforms across various LiDAR FoV configurations. The key contributions of this paper are:

- **Enhanced RRT-Based Frontier Point Search:** We employ the widely approved local-global exploration architecture based on multiple RRTs like [8], [10], and [23]. Compared to [8], our approach introduces robot traversability checks during RRT expansion. Inspired by [23], we implement global RRT pruning upon map updates, eliminating unreachable or invalid frontier points. In contrast to [10], our sampling density adjustment is based on obstacle distribution rather than the known area, improving exploration potential in complex and constrained environments.
- **Stepwise and Consistent Motion Pattern Towards Precise Mapping:** To mitigate the adverse effects of messy robot motion patterns on SLAM in computing resource-constrained and LiDAR FoV-constrained robots, we use collision-free polyline paths with uniform line segment lengths to guide robot motion in exploration-based mapping tasks. The robot follows straight-line segments and performs in-place rotations at junctions to align with the next segment until frontier point navigation completed. Since in-place rotations contribute more to map expansion, polyline paths should maximize cumulative rotation angles of the robot at segment junctions while maintaining uniform segment lengths. To achieve this, we propose a polyline path planner that integrates Dijkstra's algorithm as the frontend and both Bidirectional Interpolation (BI) and Dynamic Programming (DP) as backend. Compared to path simplification methods [16] and [17], our approach is more suited for guiding robot motion in exploration-based mapping.
- **Framework Design with Process Completeness Guarantee and Modern SLAM Applicability:** The proposed framework establishes a closed-loop process from the beginning of exploration to the completion of mapping then automatically returns to the starting point. To prevent robot oscillations from frequent goal switching, execution of frontier point selection and navigation are serialized. Moreover, visited waypoint retracing is introduced, enabling repeated generating observations to promote SLAM loop closure detection and backend optimization.

## IV. Framework Design

**Figure 5** shows the overall architecture and process design of the proposed framework. Upon initiation, an independent global exploration thread is launched, where the global RRT expands from the starting point and continuously updates the global frontier point set $F_G$, compensating for potential omissions in local exploration. This thread also monitors map updates, pruning $F_G$ and corresponding global RRT structure when certain frontier points become unreachable. The robot first conducts a stationary local frontier point search, where the local RRT expands from robot's current pose to generate the local frontier point set $F_L$. Before each local frontier search, a 360° in-place rotation is performed to maximize the update of map known regions, improving exploration efficiency. If a feasible local frontier point is identified, the robot navigates towards it using the stepwise and consistent motion pattern until either reaching the goal or determining its infeasibility. During navigation, segment junctions visited by the robot are recorded in $P_{VISITED}$ for subsequent waypoint retracing. This process loops between local exploration and navigation. If no local frontiers exist, the best global frontier point from $F_G$ is selected. The robot then integrates the downsampled $P_{VISITED}$ and optimizes the traversal sequence using Simulated Annealing (SA) before navigating with the same stepwise and consistent motion pattern. After executing the waypoint retracing mechanism, it attempts to drive the robot to navigate the selected global frontier point. If both $F_L$ and $F_G$ are empty, exploration is deemed complete, and the robot autonomously returns to the starting pose, finalizing the mapping process.

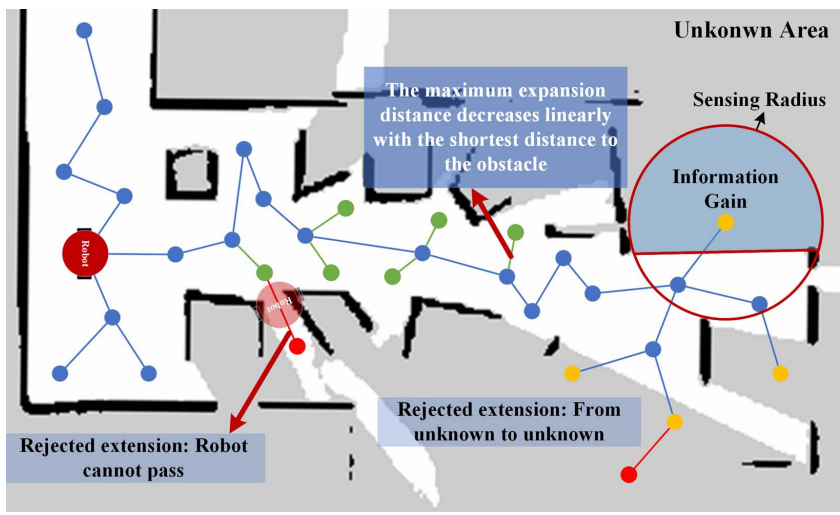

Fig. 6. Enhanced RRT frontier search and information gain calculation.

The implementation details of frontier search, frontier and RRT pruning, frontier point selection, and polyline path planning for the stepwise and consistent motion pattern are provided in the following sections.

## *A. Frontier Search and RRT Pruning*

The proposed framework utilizes an enhanced RRT for frontier search, as depicted in **Figure 6**. Unlike traditional RRT, which employs a fixed expansion step, the enhanced version dynamically adjusts the expansion step, decreasing it linearly based on the shortest distance between the node and nearby obstacles. This improves the RRT's ability to expand in narrow regions. Additionally, strict traversability checks are incorporated. An expansion is discarded if any point along the candidate path is closer to an obstacle than the robot's radius, preventing the generation of unreachable frontier points. Furthermore, expansions between unknown regions are also discarded to eliminate redundant frontiers. Finally, only nodes whose connection to their parent crosses an unknown region are considered potential frontier points. All potential frontiers are uniformly downsampled to form the final frontier point set. **Algorithm 1** presents the detailed RRT expansion and frontier point search process, employing the following functions:

***ExplorationCompleted:*** This function receives the RRT tree $\boldsymbol{T}$, the RRT sampling boundary $\boldsymbol{B}$ and the grid map $\boldsymbol{M}$, calculates the RRT extension coverage $c_{RRT}$, determines whether $c_{RRT}$ is greater than the threshold $\theta_{cov}$, and returns whether the exploration should end. For local exploration, when more than $\theta_{F_L}$ frontier points are detected, the end condition is also met. $c_{RRT}$ can be expressed as

$$c_{RRT} = \frac{N_{RRT} \cdot {r_M}^2}{S_{free}}, \quad (1)$$

where $N_{RRT}$ is the number of nodes in $\boldsymbol{T}$. $r_M$ is the resolution of $\boldsymbol{M}$. $S_{free}$ is the area of the free region of the part of $\boldsymbol{M}$ in $\boldsymbol{B}$.

***NearestNode:*** This function receives a point $\boldsymbol{p}$ and the RRT tree $\boldsymbol{T}$, returns the node in $\boldsymbol{T}$ that is closest to $\boldsymbol{p}$ in Euclidean distance.

***ExtendPoint:*** This function receives two points $\boldsymbol{p}_{rand}$ and $\boldsymbol{p}_{nearest}$ and the expansion distance $\eta$, returns new point $\boldsymbol{p}_{new}$ in the direction of $\boldsymbol{p}_{rand}$ from $\boldsymbol{p}_{nearest}$ with distance $\eta$.

***MinDistToObstacle:*** This function receives a point $\boldsymbol{p}$ and the grid map $\boldsymbol{M}$, returns the distance between $\boldsymbol{p}$ and the nearest occupied cell on $\boldsymbol{M}$ using a lookup table. The table is precomputed upon map updates, where k-d tree is utilized to determine the shortest distance from each cell to the nearest occupied cell.

***InSamplingBoundary:*** This function receives a point $\boldsymbol{p}$, returns whether the point is within the RRT sampling

**Algorithm 1** RRT Expansion and Frontier Search

**Input:** Initial root node $\boldsymbol{p}_{root}$, RRT tree $\boldsymbol{T}$, RRT sampling boundary $\boldsymbol{B}$, and grid map $\boldsymbol{M}$
**Output:** Frontier point set $\boldsymbol{F}$
**Initialize:** $\boldsymbol{T} \leftarrow \{\boldsymbol{p}_{root}\}$, $\boldsymbol{F} \leftarrow \emptyset$
**while** not $ExplorationCompleted(\boldsymbol{T}, \boldsymbol{B}, \boldsymbol{M})$ **do**
  $\boldsymbol{p}_{rand} \leftarrow Random\ point$
  $\boldsymbol{p}_{nearest} \leftarrow NearestNode(\boldsymbol{p}_{rand}, \boldsymbol{T})$
  $\eta \leftarrow AdaptiveExpandDist(\boldsymbol{p}_{nearest}, \boldsymbol{M})$
  $\boldsymbol{p}_{new} \leftarrow ExtendPoint(\boldsymbol{p}_{rand}, \boldsymbol{p}_{nearest}, \eta)$
  **if** (**not** $InSamplingBoundary(\boldsymbol{p}_{new}, \boldsymbol{B})$) **or** ($\boldsymbol{p}_{nearest}$ and $\boldsymbol{p}_{new}$ are both in unknown area) **or** (**not** $TraversabilityCheck(\boldsymbol{p}_{nearest}, \boldsymbol{p}_{new}, \boldsymbol{M})$) **then**
    **continue**
  **If** $\boldsymbol{p}_{nearest}$ is in known and $\boldsymbol{p}_{new}$ **is** in unknown area **then**
    $\boldsymbol{F} \leftarrow UniformDownsample(\boldsymbol{F} \cup \{\boldsymbol{p}_{new}\})$
  $\boldsymbol{T} \leftarrow \boldsymbol{T} \cup \{\boldsymbol{p}_{new}\}$
**return** $\boldsymbol{F}$

**Algorithm 2** Global Frontier Pruning and Global RRT Pruning

**Input:** Global frontier set $\boldsymbol{F}_G$, RRT tree $\boldsymbol{T}$, and grid map $\boldsymbol{M}$
**Output:** Pruned frontier point set $\boldsymbol{F}_G$, and pruned RRT tree $\boldsymbol{T}$
**for** each $\boldsymbol{p}_f \in \boldsymbol{F}_G$ **do**
  **if** $\boldsymbol{p}_f$ is in known area **then**
    Remove $\boldsymbol{p}_f$ from $\boldsymbol{F}_G$
    Remove corresponding node and edge from $\boldsymbol{T}$
    **continue**
  $\boldsymbol{p}_{parent} \leftarrow ParentNode(\boldsymbol{p}_f, \boldsymbol{T})$
  **while** $TraversabilityCheck(\boldsymbol{p}_f, \boldsymbol{p}_{parent}, \boldsymbol{M})$ **do**
    $\boldsymbol{p}_f \leftarrow \boldsymbol{p}_{parent}$, $\boldsymbol{p}_{parent} \leftarrow ParentNode(\boldsymbol{p}_{parent}, \boldsymbol{T})$
    **if** $\boldsymbol{p}_{parent}$ doesn't exist **then**
      **break**
  **if** $\boldsymbol{p}_{parent}$ exists **then**
    Remove $\boldsymbol{p}_f$ and all connected nodes and edges from $\boldsymbol{T}$ except $\boldsymbol{p}_{parent}$
**return** $\boldsymbol{F}_G$, $\boldsymbol{T}$

boundary $\boldsymbol{B}$. For global exploration, $\boldsymbol{B}$ is the bounding rectangle of the grid map. For local exploration, $\boldsymbol{B}$ is a circle centered at the robot position $p_{robot}$ and with the sensor observation radius $r_{sensing}$ as the radius.

***AdaptiveExpandDist:*** This function receives a point $\boldsymbol{p}$ and the grid map $\boldsymbol{M}$, returns the current adaptively adjusted maximum expansion distance $\eta$ when node $\boldsymbol{p}$ is the RRT parent node. Let $d_{obs} = MinDistToObstacle(\boldsymbol{p}, \boldsymbol{M})$ and $r_{robot}$ is the robot radius, when $d_{obs} \geq 2r_{robot}$, $\eta = \eta_{max}$, which is the maximum expansion distance of RRT. When $d_{obs} \leq r_{robot}$, $\eta = r_{robot}$.When $r_{robot} \leq d_{obs} \leq 2r_{robot}$, $\eta$ decreases linearly as $d_{obs}$ decreases.

***TraversabilityCheck:*** This function receives two points $\boldsymbol{p_0}$ and $\boldsymbol{p_1}$ and the grid map $\boldsymbol{M}$, and checks whether there is a point $\boldsymbol{p}$ on the line connecting $\boldsymbol{p_0}$ and $\boldsymbol{p_1}$ such that $MinDistToObstacle(\boldsymbol{p}, \boldsymbol{M}) < r_{robot}$. If so, it returns false, and the traversability check fails.

***UniformDownsample:*** This function receives point set $\boldsymbol{P}$, returns uniformly downsampled point set $\boldsymbol{P}_{downsampled}$.

As the global exploration RRT continuously samples and expands, some global frontier points may become explored or unreachable as the map updates. This results in invalid frontier points and unreachable nodes and edges within the global RRT. Therefore, upon each map update, global frontier pruning and RRT pruning are performed. **Algorithm 2** shows the details of the pruning process, which utilizes the following function:

***ParentNode:*** This function receives a RRT node $\boldsymbol{p}$ and the RRT tree $\boldsymbol{T}$, returns the parent node of $\boldsymbol{p}$ in $\boldsymbol{T}$.

### *B. Frontier Point Selection*

The process of selecting the best frontier point $\boldsymbol{f}_{best}$ for exploration can be expressed as

$$\begin{aligned}\boldsymbol{f}_{best} = \underset{\boldsymbol{f}\in\boldsymbol{F}}{argmax}(&w_{info} \cdot InfoGain(\boldsymbol{f}) + \\ &w_{dir} \cdot (\pi - |DirDiff(\boldsymbol{p}_{last}, \boldsymbol{p}_{robot}, \boldsymbol{f})|) + \\ &w_{dist} \cdot \|\boldsymbol{p}_{robot} - \boldsymbol{f}\|_2 + \\ &w_{free} \cdot MinDistToObstacle(\boldsymbol{f}, \boldsymbol{M})),\end{aligned} \quad (2)$$

where $\boldsymbol{F}$ is the local or global frontier point set. $w_{info}$, $w_{dir}$, $w_{dist}$ and $w_{free}$ are information gain coefficient, direction keeping coefficient, distance coefficient and area openness coefficient. $\boldsymbol{p}_{last}$ is the last visited segment junction of the robot. $\boldsymbol{p}_{robot}$ is the current position of the robot. $InfoGain(\boldsymbol{f})$ is the area of the unknown region in $\boldsymbol{M}$ covered by a circle with $\boldsymbol{f}$ as the center and $r_{sensing}$ as the radius, as shown in **Figure 6**. $DirDiff(\boldsymbol{p}_0, \boldsymbol{p}_1, \boldsymbol{p}_2)$ is the normalized angle between the line segment from $\boldsymbol{p}_0$ to $\boldsymbol{p}_1$ and the line segment from $\boldsymbol{p}_1$ to $\boldsymbol{p}_2$. This evaluation function prioritizes frontier points that are farther from the robot, aligned with the direction of the robot's trajectory, located in more open spaces, and contain a greater potential for expansion, thereby balancing efficiency, continuity, and safety.

### *C. Frontier Point Navigation*

For frontier point navigation, the robot follows a stepwise and consistent motion pattern guided by polyline global paths.

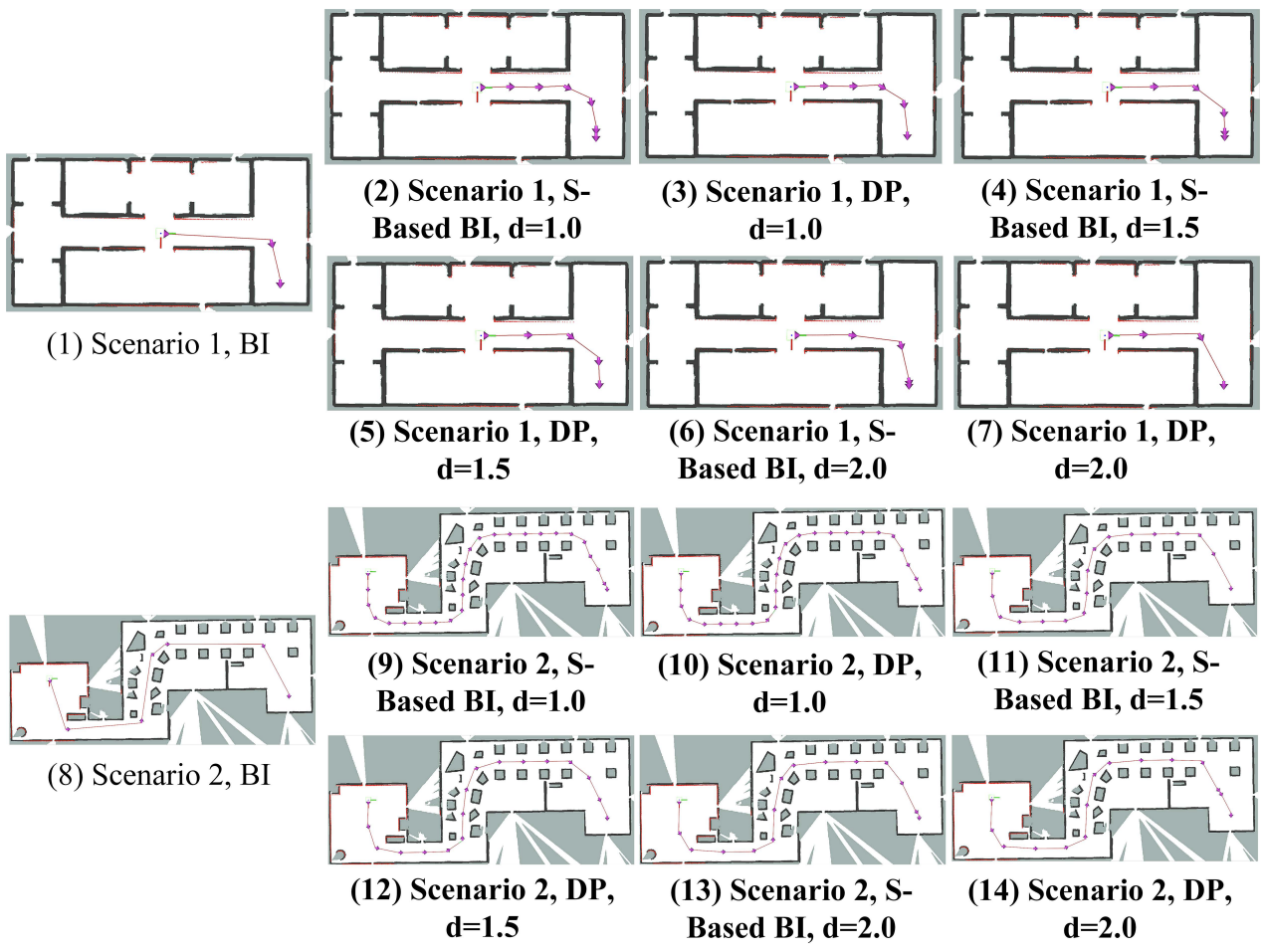

Fig. 7. Partial experimental results of the polyline path planner.

These paths should maintain uniform segment lengths while maximizing the cumulative robot in-place rotation at segment junctions. Rotations are used to align with the direction of the next segment. Once polyline global paths are generated, motion planner controls the velocity to perform high-precision straight line tracking along each segment and rotations in place at the junctions. For motion planning, this work adopts the method from [2], while for polyline path planning, we propose an approach that integrates Dijkstra's algorithm as the frontend and incorporates Bidirectional Interpolation (BI) and Dynamic Programming (DP) as the backend. The BI-based and DP-based post-processing approaches are independent, allowing either to be selected. Their respective performances will be evaluated in subsequent sections.

Let $\boldsymbol{\mathcal{X}}$ denote the initial path generated by the Dijkstra's algorithm frontend. For $\boldsymbol{x}_i \in \boldsymbol{\mathcal{X}}$, to evaluate the suitability of the exist segment from $\boldsymbol{x}_{prev} \in \boldsymbol{\mathcal{X}}$ to $\boldsymbol{x}_i$ combined with the segment from $\boldsymbol{x}_i$ to $\boldsymbol{x}_{next} \in \boldsymbol{\mathcal{X}}$ to be extended for the post-processing objective, we define the evaluation function $S(\boldsymbol{x}_{prev}, \boldsymbol{x}_i, \boldsymbol{x}_{next})$, which can be expressed as

$$\begin{aligned}S = \; &k_{rot} \cdot |DirDiff(\boldsymbol{x}_{prev}, \boldsymbol{x}_i, \boldsymbol{x}_{next})| - \\ &k_{uni} \cdot |\|\boldsymbol{x}_{next} - \boldsymbol{x}_i\|_2 - d|,\end{aligned} \quad (3)$$

where $k_{rot}$ and $k_{uni}$ are the rotation angle coefficient and length deviation penalty factor, and $d$ is the longest length of each line segment. For BI-based method, building upon [16] and [17], we refine its *Procedure for Reduce Vertices* by using $d$ as the search window and $TraversabilityCheck$ as the collision detection method, and selecting the next anchor point in a greedy manner to maximize $S$ rather than maximizing segment length. In bidirectional path selection, the reduction direction is determined by maximizing the sum of $S$ at all segment junctions rather than minimizing the number of segments. For DP-based post-processing, given an initial path $\boldsymbol{\mathcal{X}} = \{\boldsymbol{x}_0, \boldsymbol{x}_1, \dots, \boldsymbol{x}_n\}$, the goal is to determine a subset $\boldsymbol{\mathcal{X}}' = \{\boldsymbol{x}_0, \boldsymbol{x}_{i_1}, \boldsymbol{x}_{i_2}, \dots, \boldsymbol{x}_m\}$ that maximizes $\sum_{i=1}^{m-1} S(\boldsymbol{x}_{i-1}, \boldsymbol{x}_i, \boldsymbol{x}_{i+1})$. The points in $\boldsymbol{\mathcal{X}}'$ must also satisfy $\forall k \in \{1, \dots m\}$, $\left\|\boldsymbol{x}_{i_k} - \boldsymbol{x}_{i_{k-1}}\right\|_2 \leq d$ and $TraversabilityCheck(\boldsymbol{x}_{i_{k-1}}, \boldsymbol{x}_{i_k}, \boldsymbol{M})$ is true. The DP formulation recursively determines the optimal

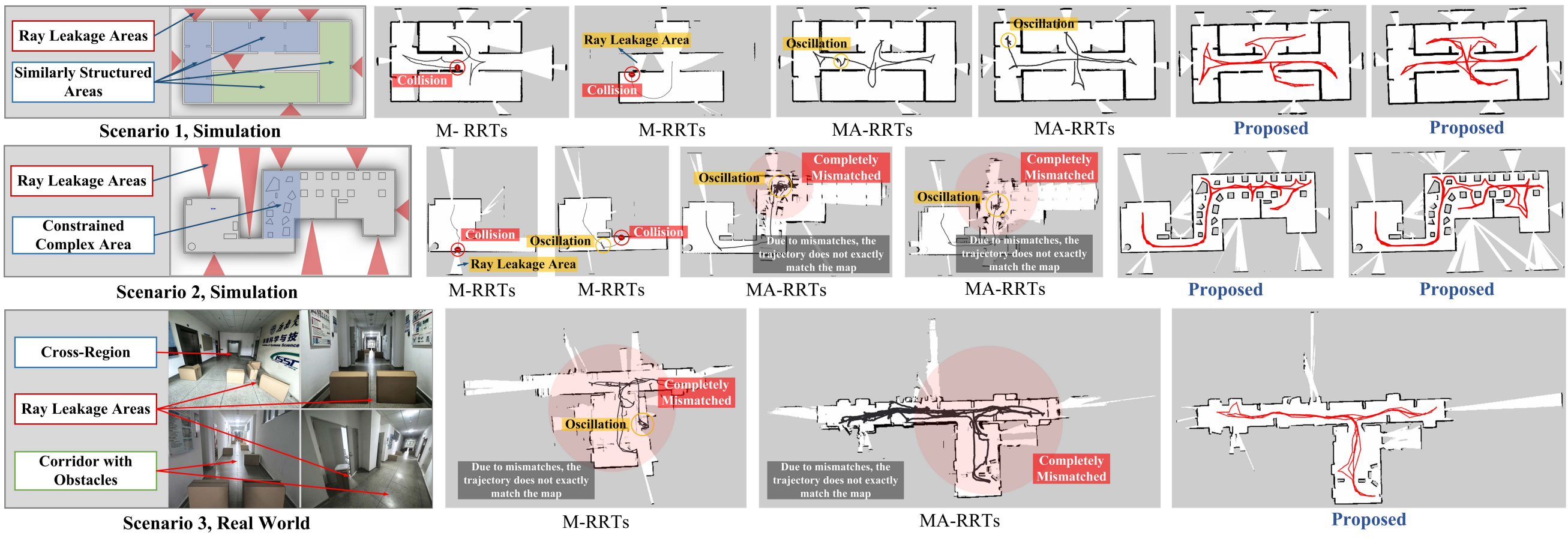

Fig. 8. Partial results of exploration-based mapping comparison experiments.

TABLE I. EVALUATIONS OF POLYLINE PATHS

| | | *S*-based BI | | | DP | | |
|---|---|---|---|---|---|---|---|
| SCN | $d$ | $s_p$ | $\varphi_{avg}^{sum}$ | $t_{avg}^{calc}$ | $s_p$ | $\varphi_{avg}^{sum}$ | $t_{avg}^{calc}$ |
| 1 | **1.0** | $0.35m$ | 1.58 | $\mathbf{0.05s}$ | $\mathbf{0.22m}$ | **1.65** | $0.16s$ |
| | **1.5** | $0.48m$ | 1.34 | $\mathbf{0.09s}$ | $\mathbf{0.34m}$ | **1.54** | $0.22s$ |
| | **2.0** | $0.57m$ | 1.11 | $\mathbf{0.12s}$ | $\mathbf{0.38m}$ | **1.36** | $0.32s$ |
| 2 | **1.0** | $0.08m$ | 6.20 | $\mathbf{0.17s}$ | $\mathbf{0.06m}$ | **6.52** | $0.53s$ |
| | **1.5** | $0.26m$ | 5.71 | $\mathbf{0.27s}$ | $\mathbf{0.11m}$ | **6.03** | $0.76s$ |
| | **2.0** | $0.50m$ | 5.67 | $\mathbf{0.36s}$ | $\mathbf{0.47m}$ | **5.88** | $1.05s$ |

TABLE II. EVALUATIONS OF EXPLORATION MAPPING APPROACHES

| | M-RRTs | | | MA-RRTs | | | Proposed | | |
|---|---|---|---|---|---|---|---|---|---|
| SCN | CO | OS | FIN | CO | OS | FIN | CO | OS | FIN |
| 1 | 8 | 6 | 6 | 4 | 6 | 10 | **0** | **0** | **20** |
| 2 | 16 | 4 | 0 | 5 | 14 | 1 | **1** | **0** | **19** |
| 3 | 7 | 12 | 1 | 3 | 13 | 4 | **0** | **0** | **20** |

path selection using the recurrence relation, which can be express as

$$DP(i,j) = \max_{k<j}\left[DP(k,i) + S(\boldsymbol{x}_k, \boldsymbol{x}_i, \boldsymbol{x}_j)\right], \tag{4}$$

where $DP(i,j)$ denotes the optimal cumulative evaluation function value for a path ending at $\boldsymbol{x}_j$, with $\boldsymbol{x}_i$ as its preceding point. The final polyline path $\boldsymbol{\mathcal{X}}'$ is obtained through a retracing procedure, where the last selected point is determined as

$$x_m = x_n, x_{m-1} = \underset{j}{\operatorname{argmax}} DP(j,n). \tag{5}$$

During navigation, continuous map updates may make the current path no longer collision-free. Upon reaching a segment junction and completing the in-place rotation to align the forward orientation with the next segment, the robot performs $TraversabilityCheck(\boldsymbol{p}_{robot}, \boldsymbol{p}_{next}, \boldsymbol{M})$, where $\boldsymbol{p}_{next}$ is the endpoint of the next line segment. If a predicted collision is detected, global path re-planning is triggered. If no feasible global path can be generated, it indicates that the current target frontier is unreachable. In local frontier point navigation, local exploration will be re-performed. During global frontier point navigation with waypoint retracing, an attempt will be made to skip the current waypoint and plan to the next waypoint. If all waypoints and the selected frontier point are unreachable, local exploration will also be re-performed.

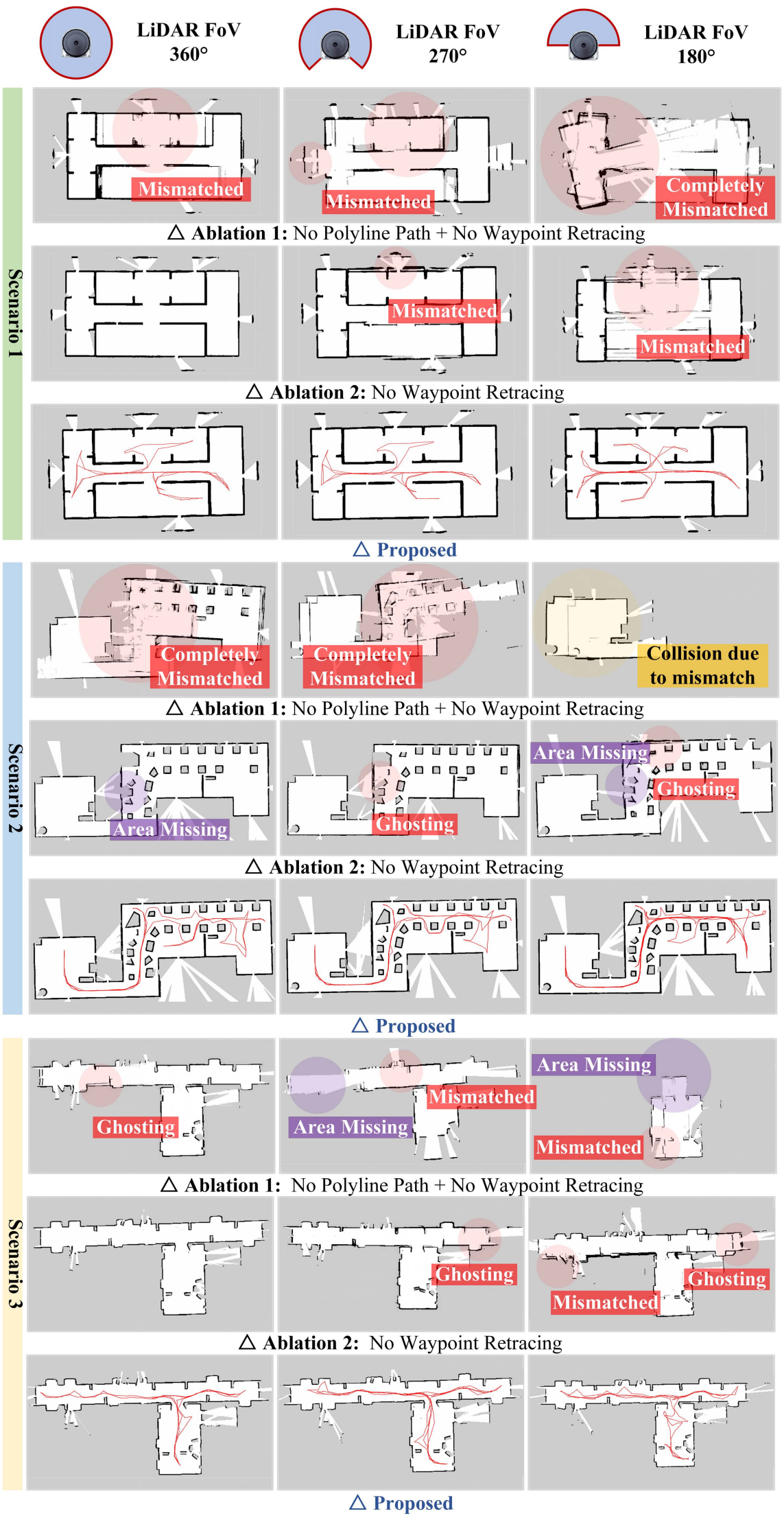

Fig. 9. Partial results of ablation experiments.

## V. EXPERIMENTS

**Figure 8** presents three sets of challenging test scenarios. The first two sets are simulation scenarios with areas of about $55m^2$ and $135m^2$, while the last set a real-world scenario with the area of about $150m^2$. Except for the Gazebo simulation running independently on a high-performance workstation, all exploration and mapping calculations are performed on resource-constrained edge computing platform based on Intel N100 (with only 4 E-cores) and ROS Noetic. The differential drive robot used in real-world experiments is shown in **Figure 5**. Its footprint is simplified to a circumscribed circle with $r_{robot} = 0.24m$. It has a 2D LiDAR with $r_{sensing} = 20m$. For mapping, *Cartographer* is used, which is a graph-based SLAM system. To plan and explore on occupancy probability maps it outputs, original maps are ternary processed into traditional occupancy grid maps with three distinct states: unknown, free, and occupied, before being fed into subsequent algorithms. The conducted experiments include: (1) Polyline path planning experiments. (2) Exploration-based mapping comparison experiments. (3) Ablation experiments with multi-FoV configurations. (4) Mapping time and finish rate experiments (5) Repeated navigation experiments.

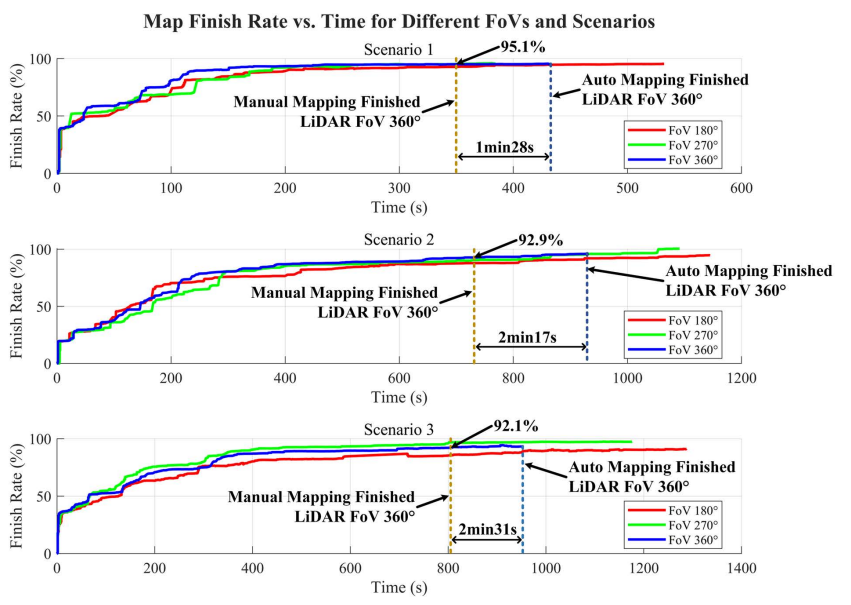

Fig. 10. Results of mapping time and convergence experiments

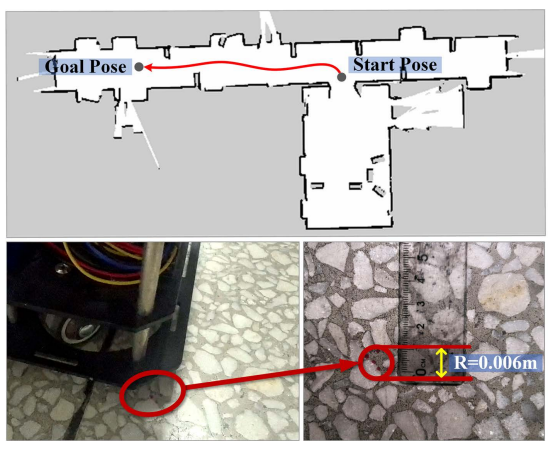

Fig. 11. Results of repeated navigation experiments.

### A. Polyline Path Planning Experiments

Since unknown regions are treated as free space in path planning, and the robot remains stationary during polyline path planning based on process design, the effectiveness of polyline path planning can be equivalently validated using fully known maps. The complete maps of scenarios 1 and 2 in **Figure 8** are utilized to evaluate the proposed polyline planner, comparing $S$-Based BI (BI with modified evaluation) and DP-based post-processing. 20 experiments are conducted. **Table I** presents results for different values of $d$ ($d = 1.0$, $d = 1.5$, and $d = 2.0$), including the pooled STD $s_p$ of segment lengths, the average cumulative in-place rotation angle $\varphi_{avg}^{sum}$, and the average computation time $t_{avg}^{calc}$ for both methods. **Figure 7** further illustrates planned polyline paths of original BI, $S$-Based BI, and DP. Overall, original BI is unsuitable, while DP achieves more globally optimal solutions but requires higher computation times than $S$-Based BI. Based on data analysis, both DP and $S$-Based BI are practical and have similar effects. They will be randomly selected in subsequent experiments. Moreover, to balance efficiency and observation density, $d$ is set to 1.25.

### B. Exploration-Based Mapping Comparison Experiments

In these experiments, the proposed algorithm is compared with exploration based on Multiple RRTs (M-RRTs) [8] and Multiple Adaptive RRTs (MA-RRTs), which incorporates the adaptive RRT expansion mechanism from [10] into [8]. The LiDAR FoV is 360°. $\eta_{max}$ of all methods is set to $0.5m$. The proposed framework uses $\theta_{cov} = 0.95$, $\theta_{F_L} = 5$, a maximum linear velocity of $1.25m/s$, and a maximum angular velocity of $1.57rad/s$. M-RRTs and MA-RRTs use a maximum linear velocity of $0.5m/s$, roughly equivalent to the average velocity of the proposed framework, and a maximum angular velocity of $1.57rad/s$. Each approach is run 20 times in each scenario. **Table II** presents the number of finishes (FIN), and failures because of collisions (CO) or oscillation-caused deadlocks (OS). The scenarios involve challenges such as ray leakages and constrained areas. For baselines, finish is defined as the map being visually complete. The proposed framework performs better in autonomous mapping tasks in all three scenarios, without collision, oscillation, or deadlock. **Figure 8** illustrates sample final maps, showing that MA-RRTs outperformed M-RRTs, while the proposed framework built the most complete, structured, and precise maps.

### C. Ablation Experiments with Multi-FoV Configurations

Ablation experiments evaluate the effectiveness of key components in the proposed framework under different robot LiDAR FoV configurations (360°, 270°, and 180°). Two ablation groups are tested: *Ablation 1* disables both polyline planner and waypoint retracing, while *Ablation 2* retains polyline planner but disables waypoint retracing. Each for 20 times. The partial results in **Figure 9** illustrate the increasing importance of the stepwise and consistent motion pattern and waypoint retracing mechanism as the LiDAR FoV decreases. Additionally, the proposed framework maintains consistent mapping results across various LiDAR FoV configurations.

### D. Mapping Time and Finish Rate Experiments

**Figure 10** illustrates the variation in mapping finish rate over time across nine autonomous exploration-based mapping trials, selected from above repeat experiments, performed in three different scenarios and with three distinct LiDAR FoV configurations. The finish rate is defined as the ratio of the known area in the current map during exploration to the known area in the fully constructed map created by a skilled human operator through manual remote control. The results demonstrate that the proposed framework consistently achieves high final finish rates across various scenarios and FoV settings, further validating its robustness and adaptability under different FoV configurations. Additionally, in terms of mapping efficiency, the proposed framework closely matches the performance of a skilled human operator. Under identical LiDAR FoV conditions of 360°, the proposed framework can complete more than 92% of the map within the time required by a human operator to complete the mapping process.

### E. Repeated Navigation Experiments

As shown in **Figure 11**, five repeated navigation trials are conducted on the autonomously constructed map using the path planner from [3] and the motion planner from [2]. The resulting positional error remains within $\pm 0.006m$, indicating that the proposed framework generates maps with sufficient precision to support high-precision navigation tasks.

## VI. Conclusion

This paper presents an autonomous exploration framework for precise mapping. It is process-complete and can achieve a closed-loop automatic exploration and mapping process. It is suitable for working with modern graph-based SLAM. The enhanced RRT-based frontier search mechanism prevents unreachable frontiers, reducing potential collisions and deadlocks. The polyline path planner, designed to guide the stepwise and consistent robot motion pattern, improves SLAM mapping precision. The serialized frontier selection and navigation process eliminates oscillations caused by frequent goal pose switching. Additionally, the waypoint retracing mechanism facilitates revisiting previously explored points, promoting loop closure detection and backend optimization in graph-based SLAM. Experiments validate the effectiveness and the robustness of the framework across complex environments, resource-constrained platforms, and LiDAR FoV configurations.

In the future, we will continue to study the integration of semantic recognition into the mapping process to achieve fully automatic deployment of robots at application sites.